\documentclass[letterpaper, 10 pt, conference]{ieeeconf}  % Comment this line out if you need a4paper

\IEEEoverridecommandlockouts                              % This command is only needed if 
\usepackage{xcolor}

\usepackage{graphicx}
\usepackage{subfig}
\usepackage{url}
\usepackage{fancyhdr}
\fancypagestyle{withfooter}{
  
  \fancyfoot[C]{\footnotesize Accepted to the IEEE ICRA Workshop on Field Robotics 2024}
}

\title{\LARGE \bf
Development of Ultra-Portable 3D Mapping Systems for Emergency Services
}

\author{Charles Hamesse$^{1,2}$, Timothée Fréville$^{1}$, Juha Saarinen$^{3}$, Michiel Vlaminck$^{2}$, Hiep Luong$^{2}$ and Rob Haelterman$^{1}$% <-this % stops a space<-this % stops a space
\thanks{$^{1}$Department of Mathematics, Royal Military Academy, Brussels, BE}%
\thanks{$^{2}$UGent - IPI - imec, Ghent University, Ghent, BE}%
\thanks{$^{3}$Future Interactive Technologies, Turku University of Applied Sciences, Turku, FI}%
\thanks{Corresponding author: {\tt\small charles.hamesse@mil.be}}
}

\begin{document}

\maketitle
\thispagestyle{withfooter}
\pagestyle{withfooter}

%%%%%%%%%%%%%%%%%%%%%%%%%%%%%%%%%%%%%%%%%%%%%%%%%%%%%%%%%%%%%%%%%%%%%%%%%%%%%%%%
\begin{abstract}

Miniaturization of cameras and LiDAR sensors has enabled the development of wearable 3D mapping systems for emergency responders. These systems have the potential to revolutionize response capabilities by providing real-time, high-fidelity maps of dynamic and hazardous environments. We present our recent efforts towards the development of such ultra-portable 3D mapping systems. We review four different sensor configurations, either helmet-mounted or body-worn, with two different mapping algorithms that were implemented and evaluated during field trials. The paper discusses the experimental results with the aim to stimulate further discussion within the portable 3D mapping research community.
\end{abstract}

%%%%%%%%%%%%%%%%%%%%%%%%%%%%%%%%%%%%%%%%%%%%%%%%%%%%%%%%%%%%%%%%%%%%%%%%%%%%%%%%
\section{INTRODUCTION} 
Recent technological developments in the fields of machine vision and 3D perception sensors show immense promise for revolutionizing emergency response capabilities. The miniaturization and cost reduction of cameras and Light Detection and Ranging (LiDAR) sensors have opened doors for the development of ultra-portable 3D mapping systems, which could be contained on a setup small and light enough to be mounted on a security helmet or worn on a tactical chest rig. Nowadays, commercial off-the-shelf solid-state LiDARs can weigh less than 300 grams and be purchased for less than a thousand USD \cite{livox-mid360}, and cameras can be found weighing only a few grams in a package smaller than a die \cite{ximea}. These technologies have the potential to be a game-changer for emergency service units, special forces operators, search and rescue (SAR) personnel and firefighters: they must often operate in a highly coordinated manner, in dynamic and hazardous places where basic descriptions (e.g., pictures and text) are largely insufficient to communicate a clear spatial representation of the environment. Moreover, time, budget or space constraints will often make the use of classical 3D mapping tools (e.g., survey LiDAR) difficult. A portable 3D mapping system that can create real-time, high-fidelity representations of the scene, including collapsed buildings, narrow passages, cluttered spaces, or uneven terrain and be helmet-mounted or worn would allow for:
\begin{itemize}
    \item Improved situational awareness: first responders can visualize the entire environment, pinpoint potential hazards, and plan safer routes for themselves and potential victims.
    \item Enhanced coordination and communication: a shared 3D map between local teams and remote command posts provides a common reference point for planning rescue strategies.
    \item Faster decision making: 3D maps can expedite critical decision-making by offering a clear picture of the situation, which can be vital in time-sensitive operations.
\end{itemize}

While portable systems like NavVis \cite{navvis}, GeoSLAM \cite{geoslam}, Leica BLK2GO \cite{leicablk} exist for rapid mapping purposes, they often require at least one hand to operate and remain relatively heavy and expensive (typically costing tens of thousands of dollars). Unmanned aerial vehicles (UAVs) are another option for fast mapping, but their use can be restricted by legal regulations or environmental limitations (e.g. narrow passages with low visibility), in addition to constraints related to acquisition cost, range or endurance.

\begin{figure}[t]%
    \centering
    \includegraphics[width=8.8cm]{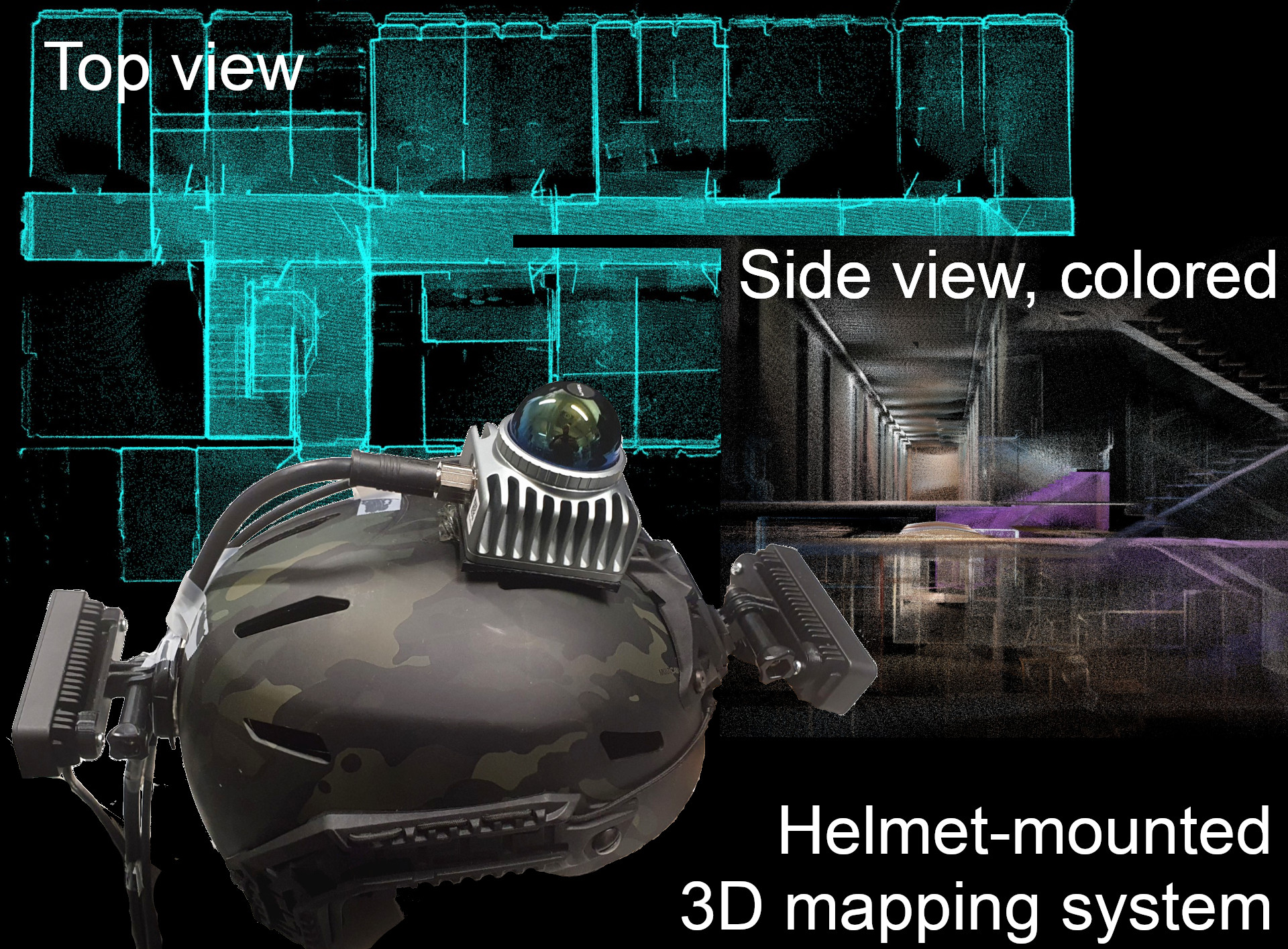}
    \caption{Illustration of one of our mapping systems, showing example result point clouds and the prototypical hardware system.}
    \label{fig:intro}%
\end{figure}

In this brief paper, we describe our past and ongoing research efforts and experimental results related to the development of ultra-portable 3D mapping systems. More precisely, we review four different mapping systems with sensors either helmet-mounted or body-worn and various sensor fusion algorithms that we have implemented and evaluated during various field trials. We discuss our experimental results, giving actionable insights and comments, in the hope to stimulate further discussion within the portable 3D mapping community.

\section{RELATED WORK}
We start with a brief review of recent sensors relevant for portable 3D mapping systems, then move on to describe the current state-of-the-art algorithms working with such hardware.

\subsection{Portable sensors} Common portable 3D mapping systems principally rely on LiDAR sensors, to which we usually add an Inertial Measurement Units (IMU), which can help making the pose estimation more robust, and a camera to perform visual feature tracking (visual odometry), and further increase the precision and robustness. 
\begin{itemize}
    \item LiDAR: following the classical spinning LiDAR devices such as the Velodyne / Ouster family  \cite{ouster}, smaller, hybrid and solid-state LiDARs are becoming the norm in portable SLAM applications, with devices such as the Livox Avia \cite{livox-avia}, Livox Mid-360  \cite{livox-mid360} or Blickfeld Qb2 \cite{blickfeld}. Most recent devices have either a conical or a cylindrical scan pattern, weigh less than 500g and have a power consumption below 15W. Although  they are different technologies in practice, we also put the time-of-flight (ToF) cameras in this category, since they also capture 3D point clouds. These cameras, such as the Intel Realsense L515 \cite{realsense} or Microsoft Azure Kinect DK \cite{kinect}, rely on low-power infrared signals and cannot be used reliably in outdoor environments due to interference with sun rays. 
    \item IMU: first, let us note that most LiDAR manufacturers embed an IMU in LiDAR devices, which is extremely helpful in practice: extrinsic calibration can be known (or at least estimated) in advance, and time synchronization between LiDAR and IMU is ensured at the hardware-level. This is the case for the Livox range and most Ouster devices, for example. External IMU modules also exist and can be desired if one wants specific characteristics or increased accuracy, e.g. with the Xsens line \cite{xsens}. That being said, our use case requiring to reduce the form factor and weight of the entire system as much as possible, we rely on the embedded IMUs of LiDAR devices.
    \item Camera: for all applications related to sensor fusion and 3D mapping, arguably the most important characteristics for cameras are a global shutter (all pixels of each image are captured at the same instant) and the possibility of hardware-synchronization (trigger pin, to make timestamps consistent with the LiDAR-inertial system). For visual feature tracking and odometry on a system moved by a human operator, a frame rate of 30 Hz and a megapixel resolution is more than enough; and those characteristics can easily be obtained with any recent camera \cite{alliedvision} \cite{flirgrass} \cite{ximea}. Then, depending on the applications, one will prefer large focal length lenses (smaller field of view, fewer features but lower distortion) or short focal length lenses (larger field of view, more features visible in each image but risk of greater distortion). Similarly to LiDAR, camera manufacturers acting in the field of robotics sometimes embed an IMU with camera sensors \cite{luxonis} \cite{realsense}. 
\end{itemize}

\subsection{Algorithms} Although the hardware already allows the development of prototypes of such systems, there remains a number of practical challenges. The algorithms for Simultaneous Localization and Mapping (SLAM) used in this context primarily consist of a LiDAR-inertial odometry (LIO) \cite{slict} \cite{Xu2021FASTLIO2FD} \cite{Bai2022FasterLIOLT}, visual-inertial odometry (VIO) \cite{Qin2017VINSMonoAR} \cite{MurArtal2015ORBSLAMAV} \cite{Geneva2020OpenVINSAR}, or LiDAR-visual-inertial odometry (LVIO) \cite{Lin2021R3LIVEAR} \cite{Lv2021CLINSCT} system, to which a loop closure method is usually added to refine the map as the user visits the same place multiple times \cite{Vlaminck2018HaveIS} \cite{Kim2018ScanCE}. In fact, state-of-the-art LIO, VIO or LVIO methods can still fail in some particularly hard cases or \textit{degenerate} scenarios. Depending on the sensor, a degenerate scenario can be different things: LiDAR registration will suffer in the absence of sufficient geometrical features (e.g., conical pattern LiDAR pointing to the floor), and visual tracking will suffer in the absence of robust features (e.g., only flat walls in the image). Unfortunately, SAR operations will often happen in degenerate environments featuring narrow passages, compact staircases and flat walls. Therefore, one could think that a multi-sensor system is always preferred, and that visual information will nicely complement the LiDAR point clouds. However, in practice, this is not always the case since these multi-sensor systems require extremely precise calibration (which is a research topic on its own \cite{Qin2018OnlineTC} \cite{Koide2023GeneralST} \cite{Rehder2016ExtendingKC}) and rigidly attached sensors to work properly. Also, they are generally more complex systems, hence more difficult to implement, harder to maintain and more expensive. % we field trials were carried out using a custom LVIO method based on loose coupling: a separate VIO thread feeds the relative pose estimates to a LIO algorithm, which incorporates these relative pose observations in the odometry optimization method, next to the scan-to-map registration residuals. 

\section{PROPOSED SYSTEMS}
We present four systems in this paper. Experiments A, B and C focus on the performance of a given LVIO method with various sensor setups in realistic, complex environments. For those, we use a custom LVIO method which uses an error-state iterative Kalman filter (ESIKF) to fuse IMU-propagated state estimates with i) point-to-plane distance residuals, in a manner similar to Fast-LIO \cite{Xu2021FASTLIO2FD} and ii) visual-inertial relative pose residuals, which we compute based on the output poses from VINS-Mono \cite{Qin2017VINSMonoAR}. The method is for the most part the same between these three experiments, with slight adjustments to cope with the different sensor configurations. (For example, LiDAR scans need to be deskewed, whereas ToF do not.) Experiment D does not use cameras, but aims to increase the coverage of a LIO system by combining two LiDAR-inertial devices in a non-rigid manner. Both devices are fixed with velcro pads on the operator's jacket. There, the ESIKF is used again, but this time for the fusion of both LIO sub-systems, sharing a common map.

\section{EXPERIMENTS}
For each experiment, we provide a hardware setup description, illustrations of the hardware prototype, visualizations of the point clouds in various test environments, and comments on the results. Our goal is to provide an overview of qualitative comments on the general outlook of the resulting 3D maps for each system. 

%%%%%
%%%%%
%%%%% A
%%%%%
%%%%%
\begin{figure}[t]%
    \centering
    \subfloat[\centering Hardware system]{{\includegraphics[width=4.5cm]{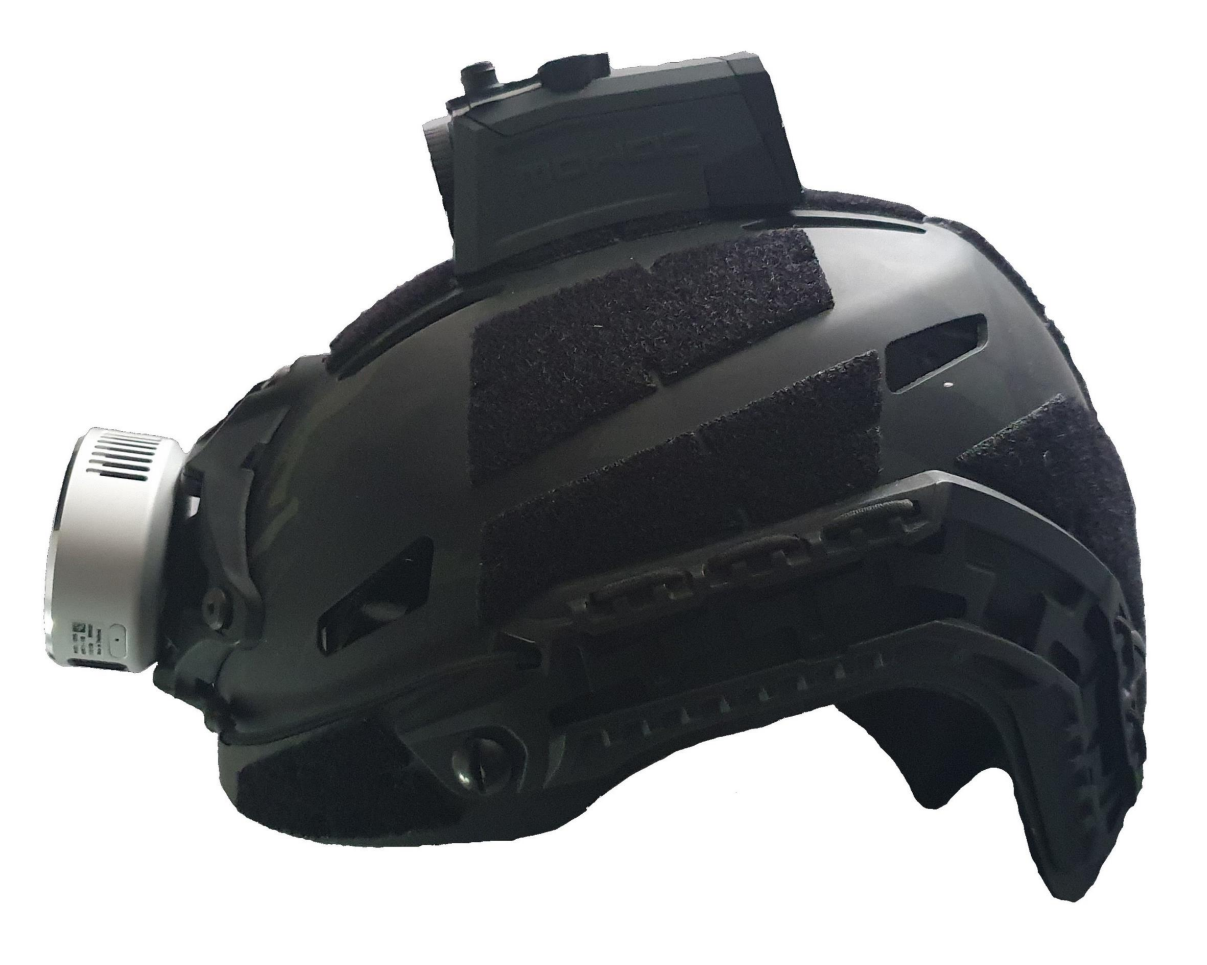} }}\\
    \subfloat[\centering Top view of the lab environment]{{\includegraphics[width=8.5cm]{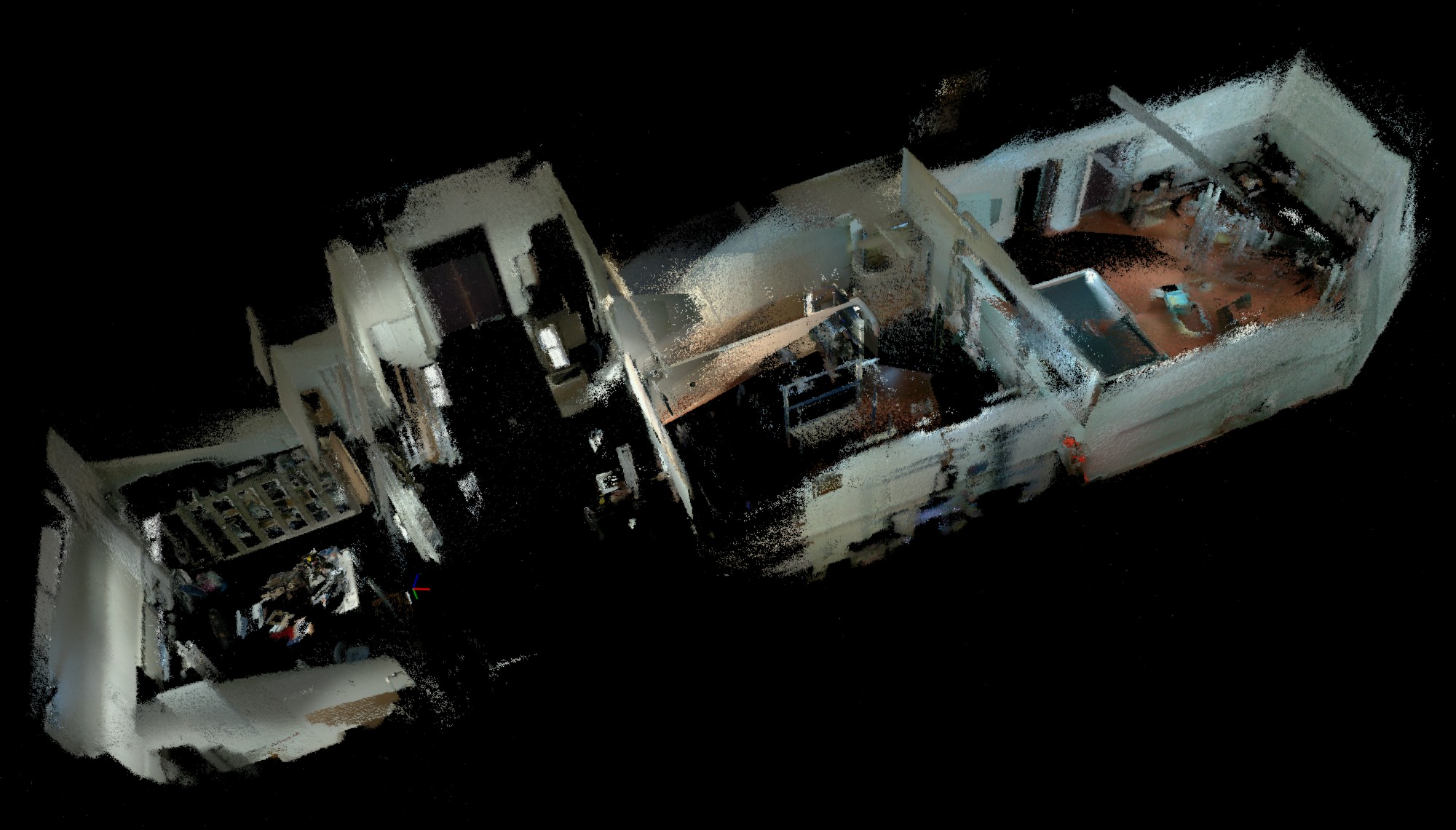} }}\\%
    \subfloat[\centering Top / side view of the office environment]{{\includegraphics[width=8.5cm]{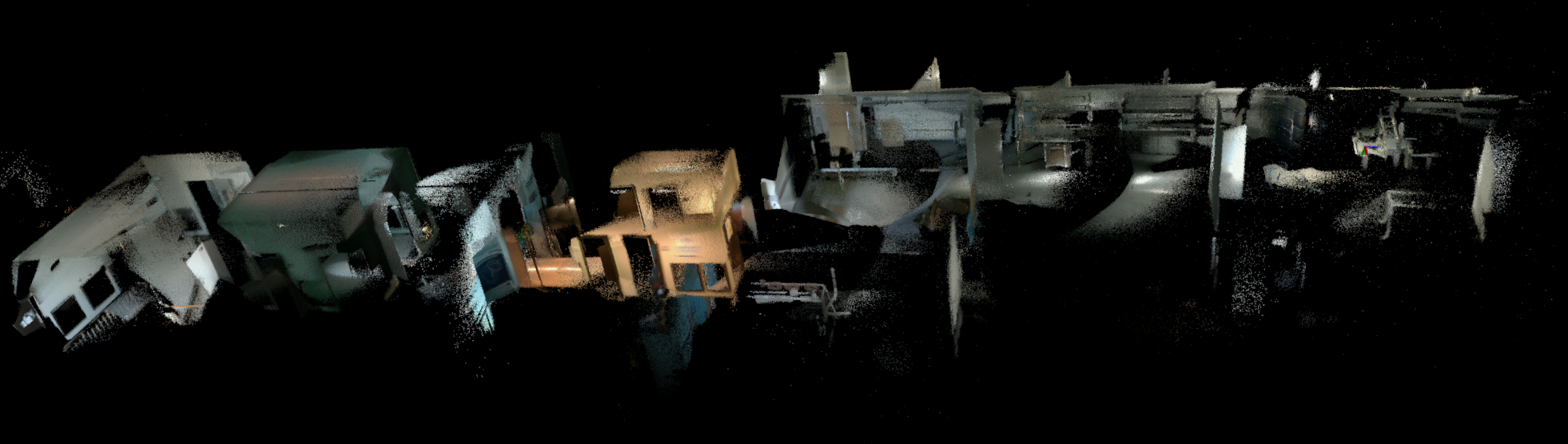} }}%
    \caption{Experiment A. Helmet-mounted Intel Realsense L515 ToF camera, including RGB camera and IMU (cable not shown). The action camera mounted on top of the helmet is not used for mapping.}%
    \label{fig:exp:L515}%
\end{figure}
\subsection{LVIO with helmet-mounted Intel Realsense L515}
\begin{figure}%
    \centering
    \subfloat[\centering Hardware system for Experiment B]{{\includegraphics[width=.28\textwidth]{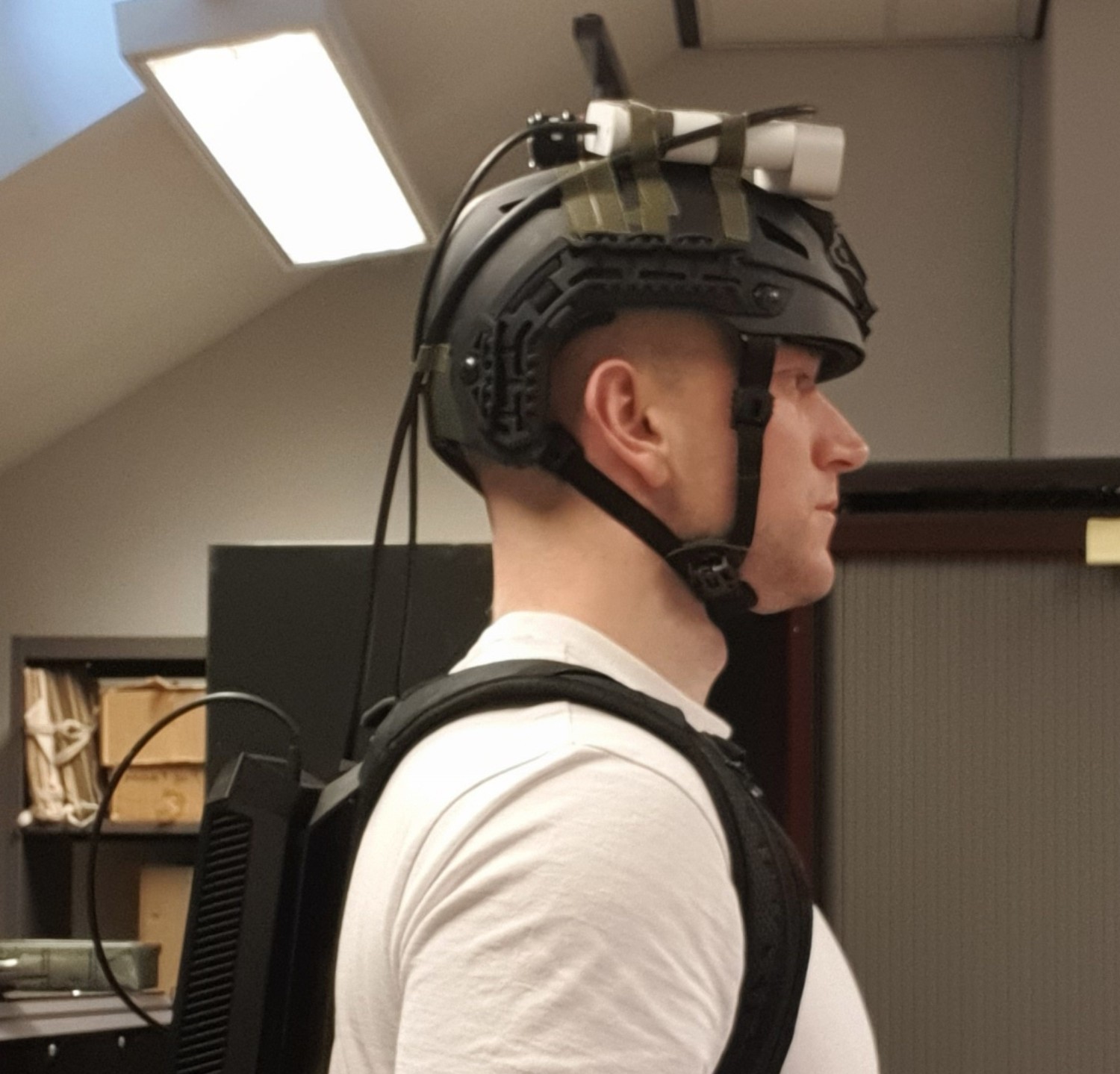} }}\\%
    \subfloat[\centering Suburban house: front view]{{\includegraphics[width=.46\textwidth]{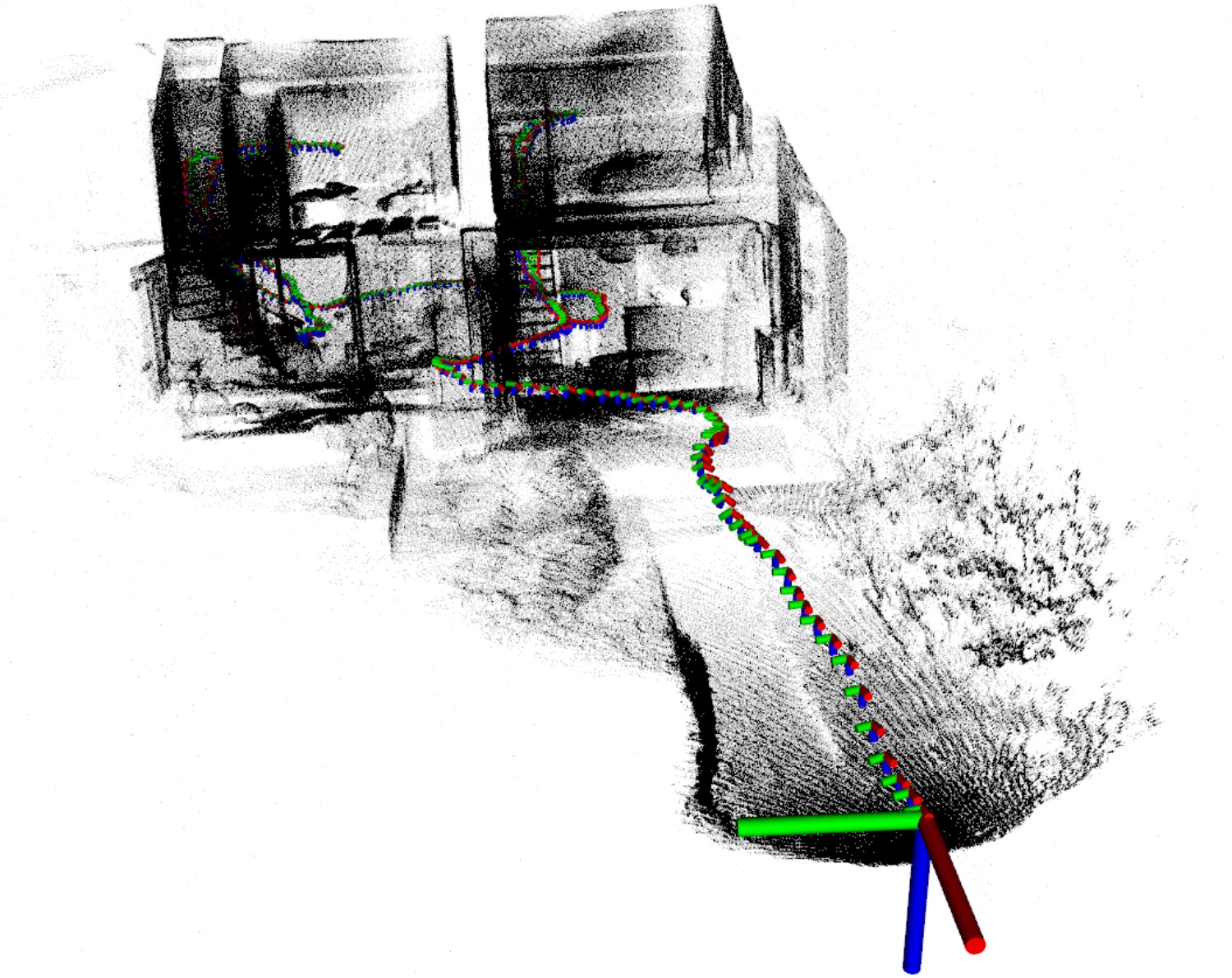} }}\\%
    \subfloat[\centering Office space]{{\includegraphics[width=.46\textwidth]{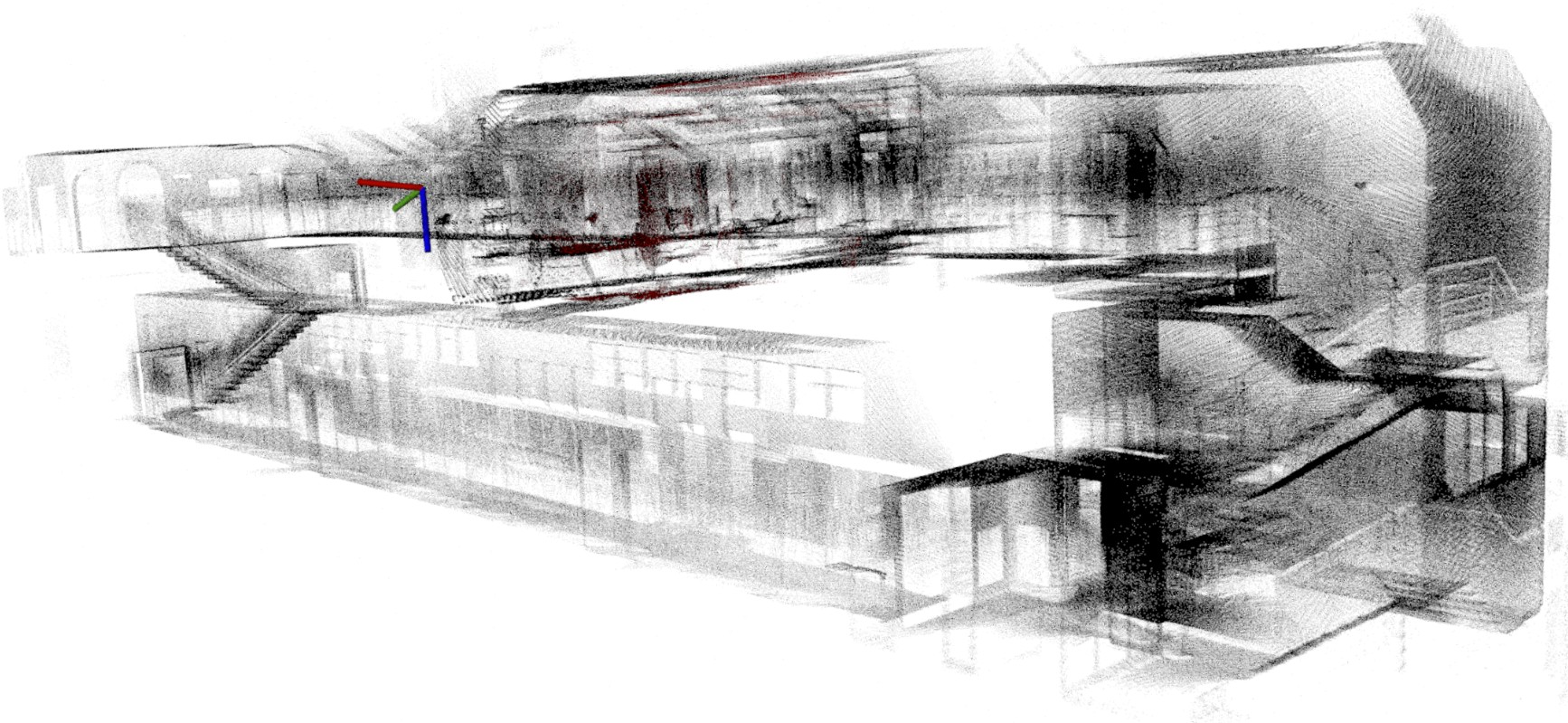} }}\\%
    \subfloat[\centering Stair case hall]{{\includegraphics[width=.24\textwidth]{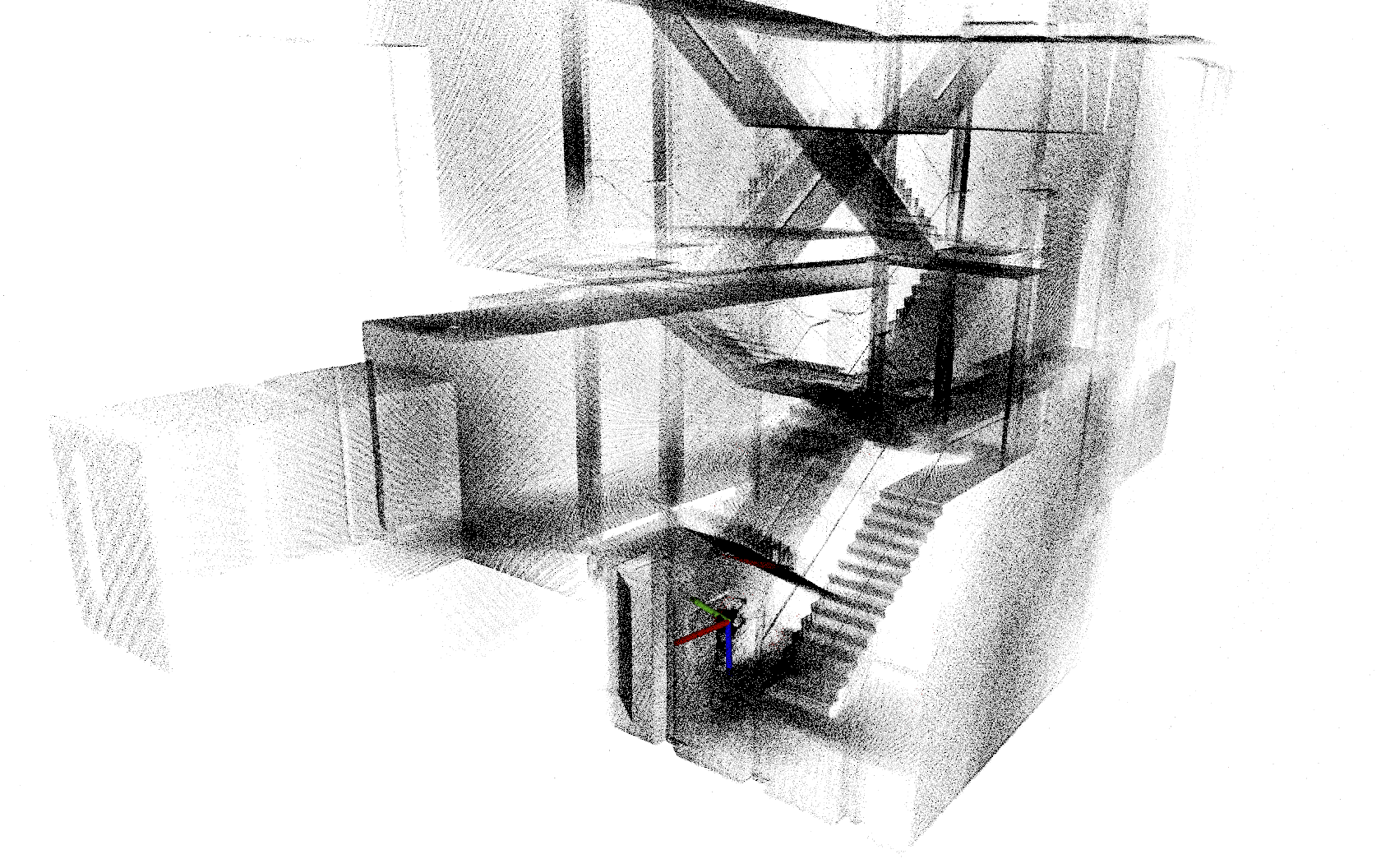} }}%
    \subfloat[\centering Semi-indoor parking lot]{{\includegraphics[width=.24\textwidth]{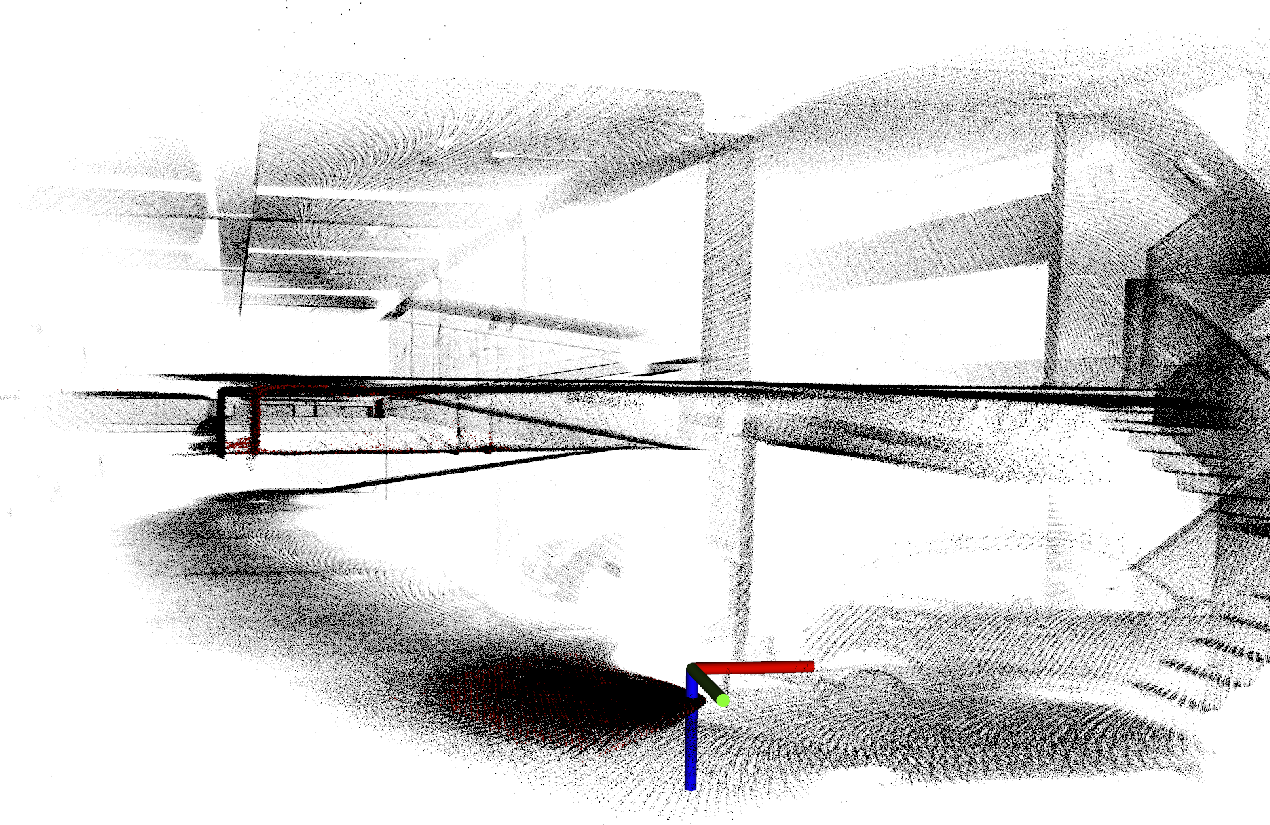} }}%
    \caption{Experiment B. Helmet-mounted Microsoft Azure Kinect ToF camera, including RGB camera and IMU, connected to a backpack PC.}%
    \label{fig:exp:kinect}%
\end{figure}
For this first experiment, we use an Intel Realsense L515 camera, mounted in the front of an OpsCore-type military helmet. We navigate in a lab environment and a large office space, taking care to perform smooth head movements. The results are shown in Figure \ref{fig:exp:L515}. Although the global layout is coherent, we can see that the sensor does not capture enough points for a complete view of the environment (the ground plane is often missing), and we can detect slight drift errors when looking closely. Of course, additional research effort on the algorithms could certainly improve these results to some extent, but the sensor already shows fundamental limitations for dense mapping, such as the limited field of view of the TOF sensor (70$^{\circ} \times 55^{\circ}$) and its low power (scan quality is degraded quickly as soon as the room there is sunlight, even through windows) and limited range (9m). Moreover, the RGB camera of the L515 uses a rolling shutter, which has a strong detrimental effect on VIO performance. In fact, when we tested this system walking and moving at a natural pace in the office space and lab environment, the system did not successfully keep a stable odometry and quicky started to diverge.

%%%%%
%%%%%
%%%%% B
%%%%%
%%%%%
\subsection{LVIO with helmet-mounted Azure Kinect DK}

We replace the Intel Realsense L515 with a Microsoft Azure Kinect DK. Arguably, mounting that device on a helmet is not ideal, as it already comes at a weight of 440g and has a form factor difficult to rigidly fix on a helmet. However, we are interested in the sensor performance at this stage, hence our experiment\footnote{In fact, the same sensors are packaged in the Microsoft HoloLens AR headset, which already has better ergonomy than our prototype system.}.

We show our experimental setup and selected map visualizations in Figure \ref{fig:exp:kinect}. We directly notice the improved geometry of the output maps, mainly thanks to the better accuracy, range and field of view of the sensor (120$^{\circ} \times 120^{\circ}$). Although the form factor of our setup is far from ideal, the mapping performance of the system would be sufficient for most emergency use cases in small, indoor environments. Moreover, with this setup, we do not have to force smooth head motion as we did in Experiment A: even with fast translations and rotations, the odometry does not drift or diverge, as long as there are enough visual and geometrical features in the field of view and in range. However, as shown in the bottom tests of Figure \ref{fig:exp:kinect} (stair case and semi-covered parking lot), the range of the sensor can quickly become limiting in larger spaces, or spaces where indirect sunlight can penetrate. In those cases, it becomes hard for the ToF sensor to gather enough valid points for odometry (the VIO sub-system can compensate to some extent, but will not solve everything) and mapping (for which there isn't any straightforward solution, as the global point cloud is what we are interested in). Moreover, this system will clearly not work in fully outdoor environments. 

\begin{figure}%
    \centering
    \subfloat[\centering Side view]{{\includegraphics[height=3.4cm]{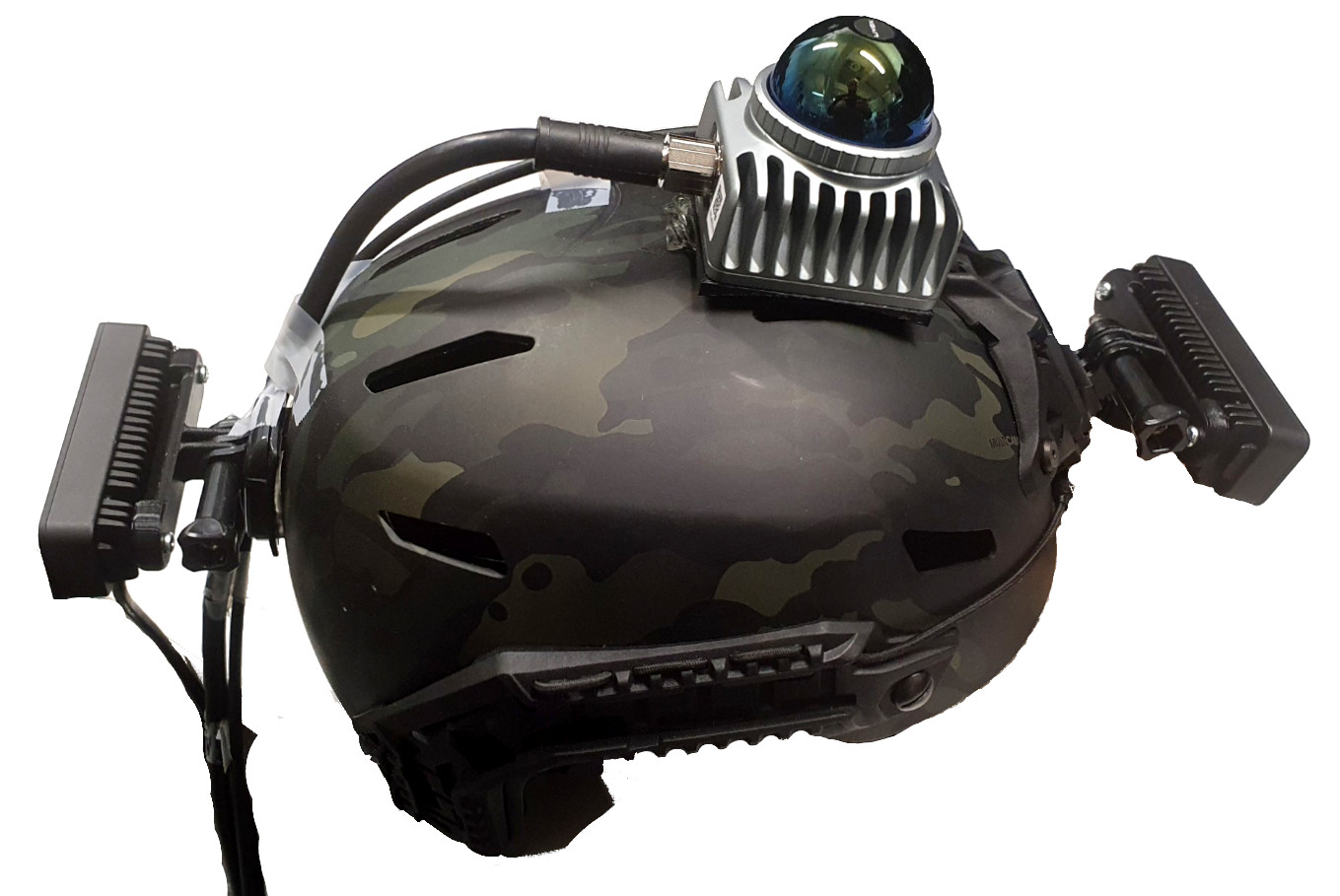} }}%
    \subfloat[\centering Front view]{{\includegraphics[height=3.4cm]{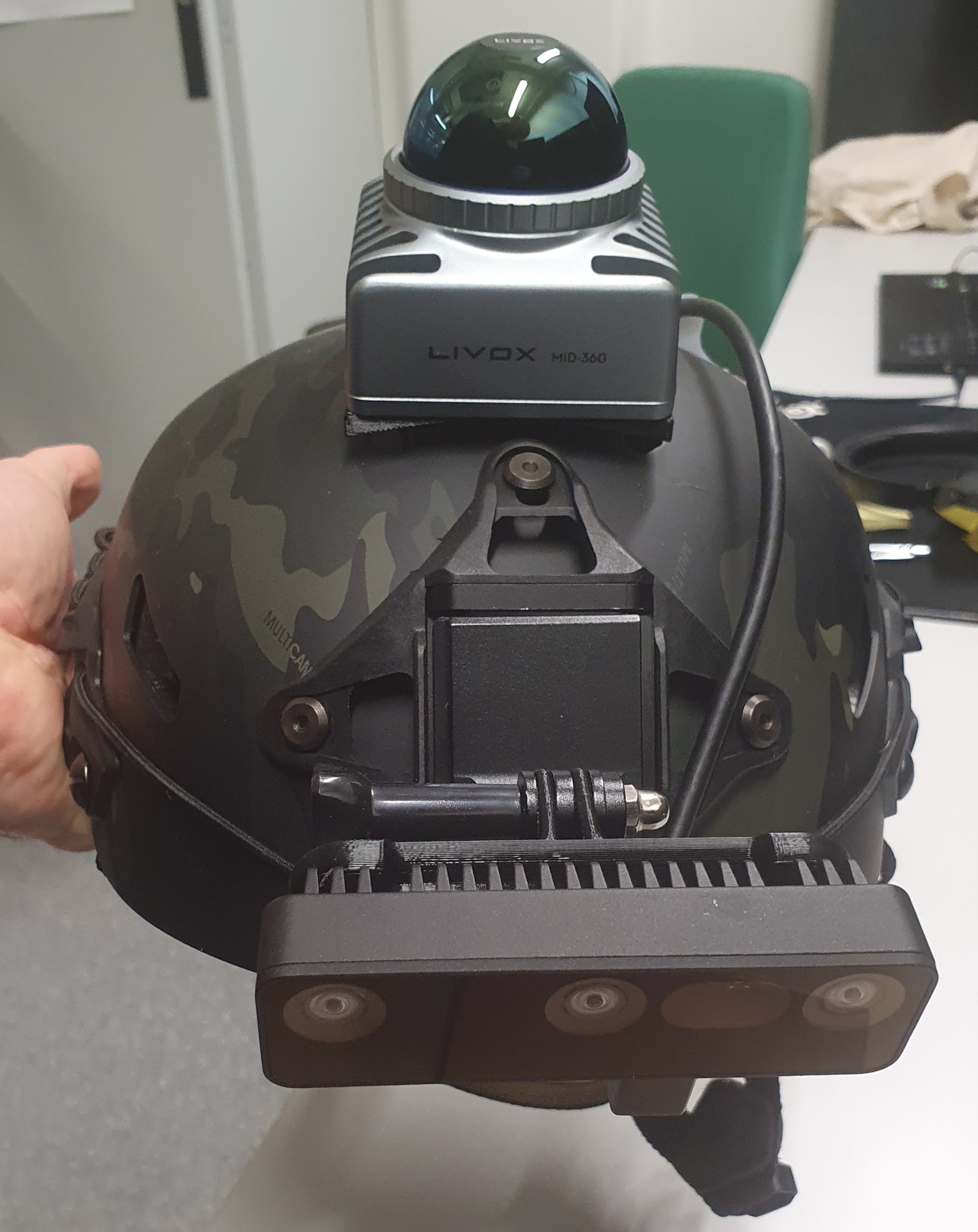} }}\\%
    \subfloat[\centering Top view]{{\includegraphics[width=.43\textwidth]{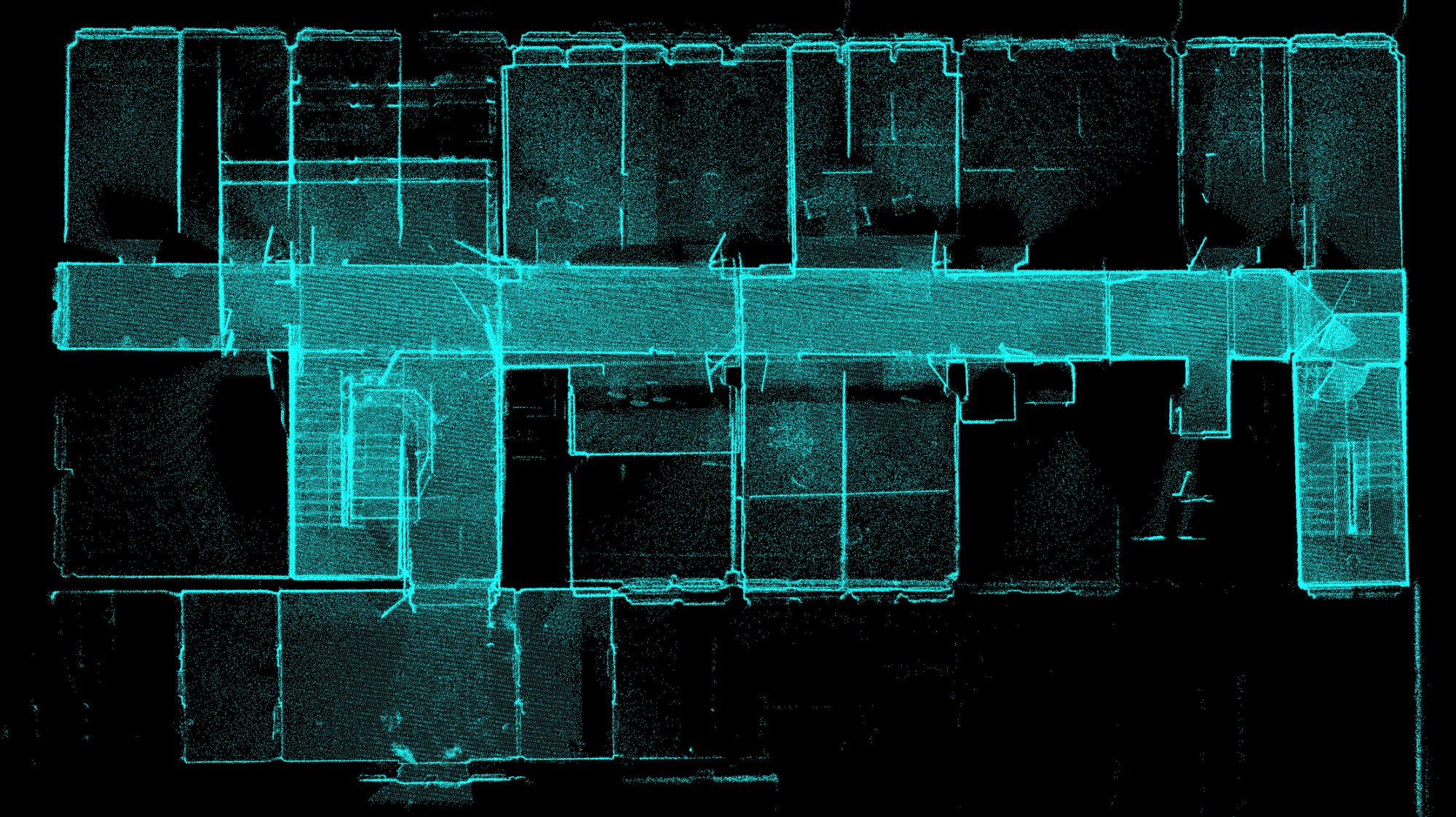} }}\\%
    \subfloat[\centering Side view 1]{{\includegraphics[width=.43\textwidth]{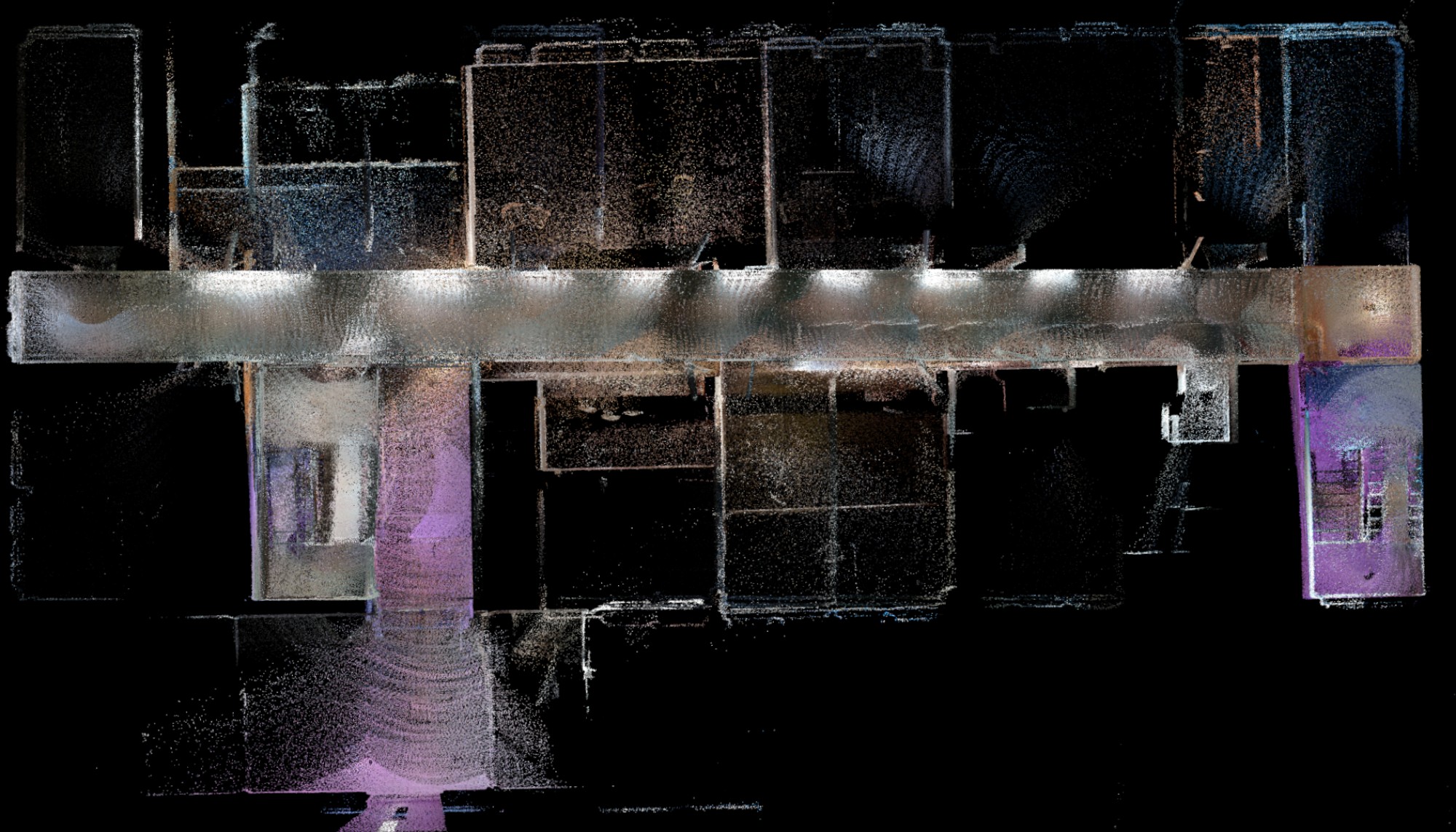} }}\\%
    \subfloat[\centering Side view 2]{{\includegraphics[width=.43\textwidth]{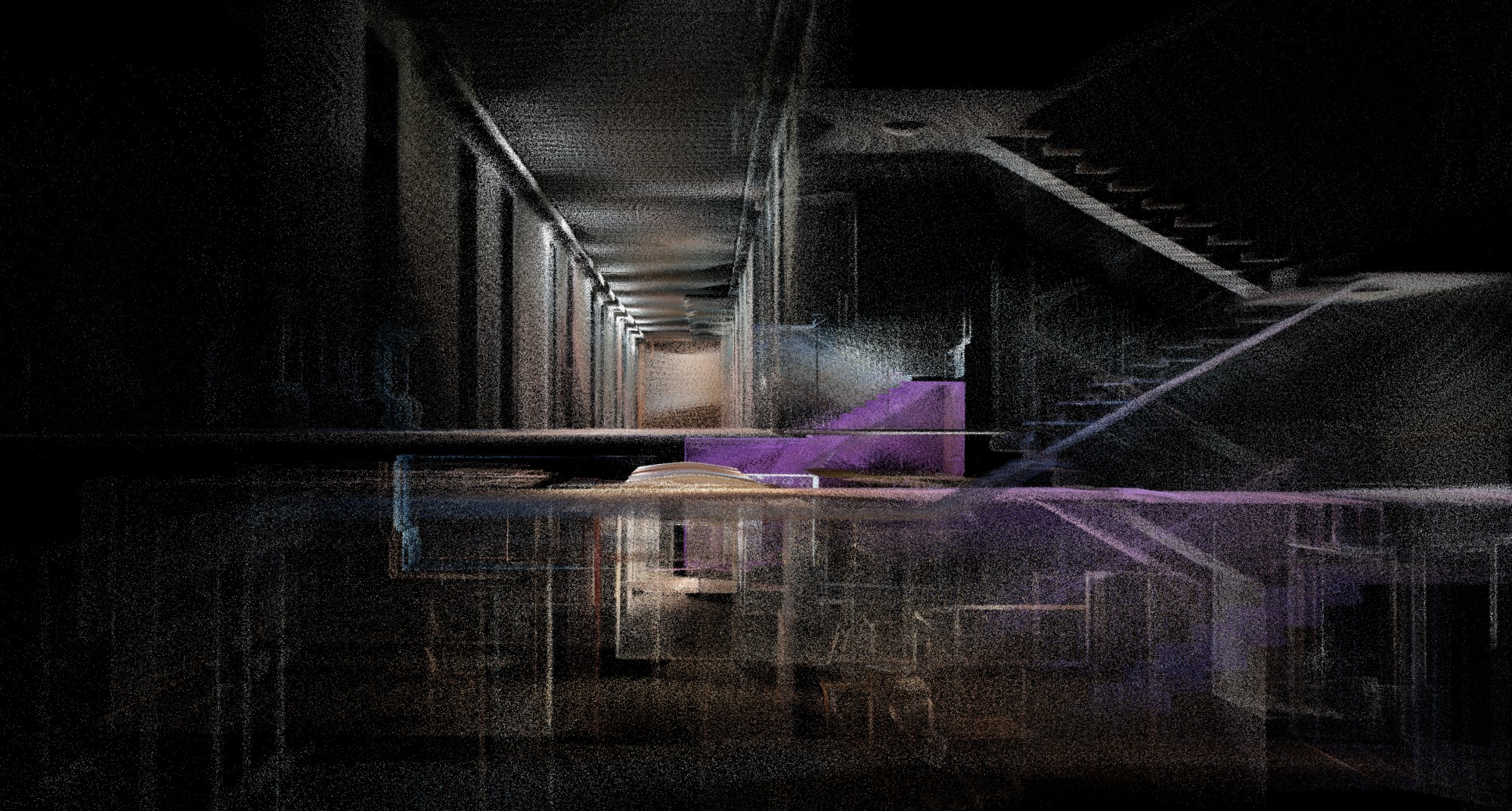} }}%
    \caption{Experiment C. Livox Mid-360 LiDAR-inertial device, with two Luxonis OAK-D Pro Wide cameras, facing backward and forward. The mapping test was realized in an office building. Colored point clouds use the RGB information of points captured by the LiDAR and visible by the cameras, which is why there are fewer points in the colored scans.}%
    \label{fig:exp:mid360oak-color}%
\end{figure}

\subsection{LVIO with helmet-mounted LiDAR-visual-inertial system with Livox Mid-360}

In this experiment, we move from a ToF camera to a small form factor, solid-state LiDAR, namely the Livox Mid-360. This allows an increased range (advertised as 40m), the ability to function outdoors, and the sensor itself weighs only 265g. However, the field of view being 360$^{\circ} \times 59^{\circ}$, the placement of the sensor in a portable system will have a crucial role in the mapping performance. (Of course, the natural head motion of the operator that occurs when walking will also have an impact of the perceived geometrical and visual features.) We consider two mounting options:
\begin{itemize}
    \item Horizontal LiDAR on top: the advantage will be that the LiDAR captures the entire horizontal layout of the room, as seen from a floor plan. This will, in theory, make the LIO system very robust as it will always have points to match all around it. However, we will generally not be able to cover the objects lower than the operator's head, which is not desirable for such mapping operations. 
    \item Tilted LiDAR (e.g., forward): there, the field of view of the LiDAR will be closer to that of the human operator, however there is an increased risk of degeneracy, especially if the operator is facing a door (most of the points that will be captured will be moving points on the door, which will lead to issues). 
\end{itemize}
Next to the LiDAR-inertial sensor, we mount two Luxonis OAK-D Pro Wide stereo visual-inertial sensors (facing backwards and forwards) to complete our LVIO sensor suite. For that reason, we choose to slightly tilt the LiDAR forward, counting on the VIO sub-system to correct potential errors made by the LIO sub-system.

\begin{figure}[h!]
    \centering
    \subfloat[\centering Illustration of the proposed dual LiDAR-inertial system (cables not shown). L1 and L2 are both attached with velcro on the tactical jacket. ]{{\includegraphics[width=.46\textwidth]{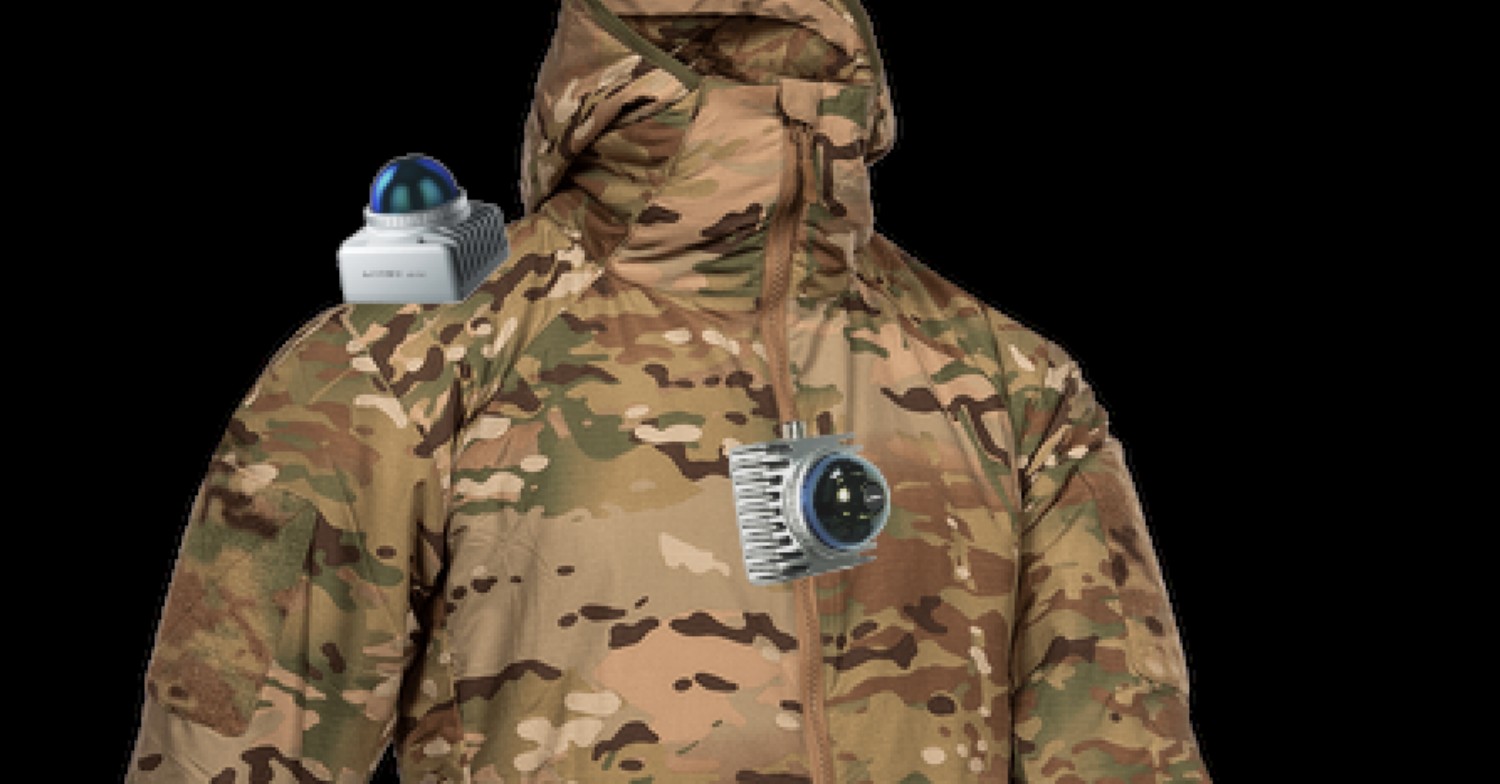} }}\\%
    \subfloat[\centering Initialization: registration of L2 into the map of L1 with point-to-plane ICP.]{{\includegraphics[width=.46\textwidth]{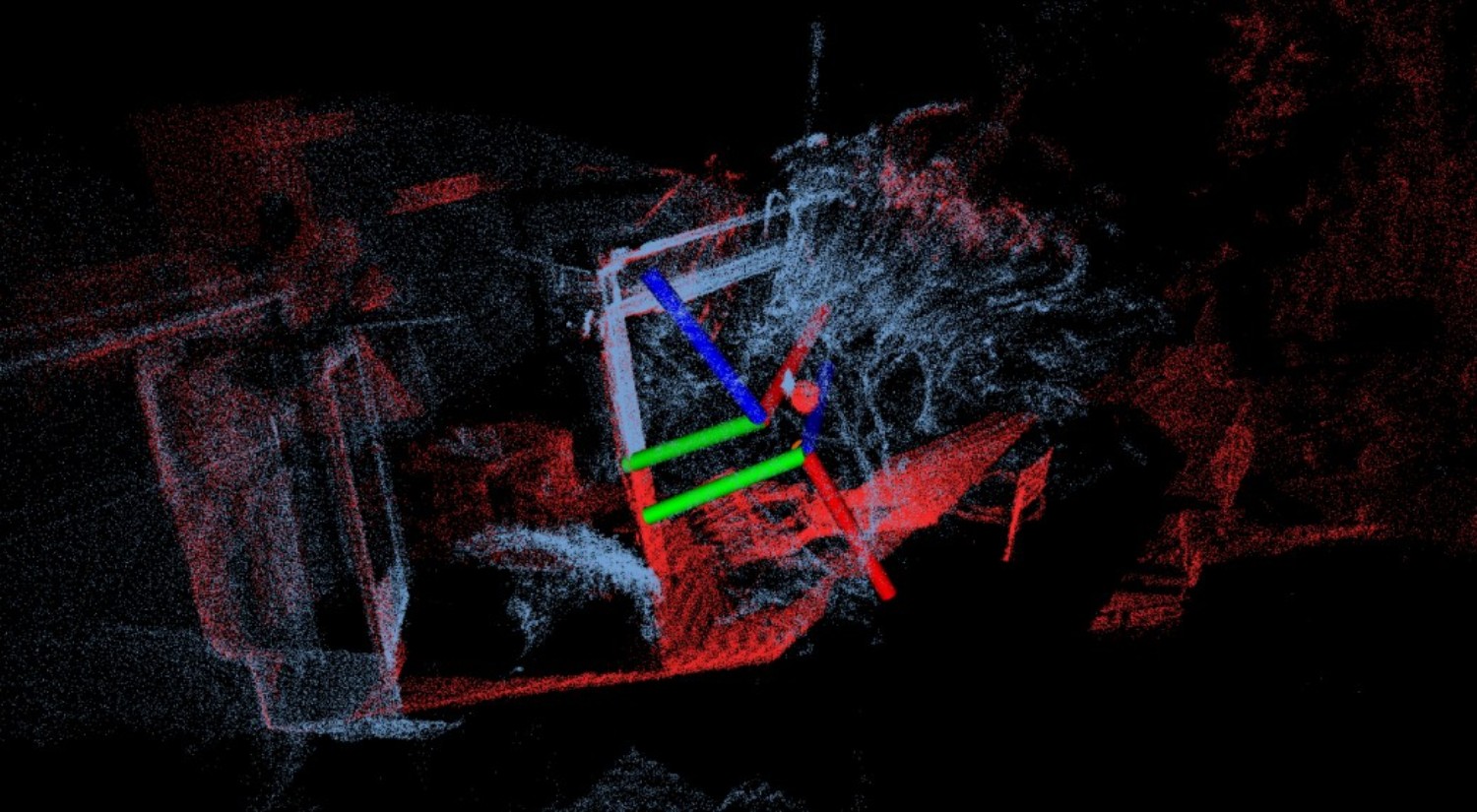} }}\\%
    \subfloat[\centering Ongoing mapping experiment.]{{\includegraphics[width=.46\textwidth]{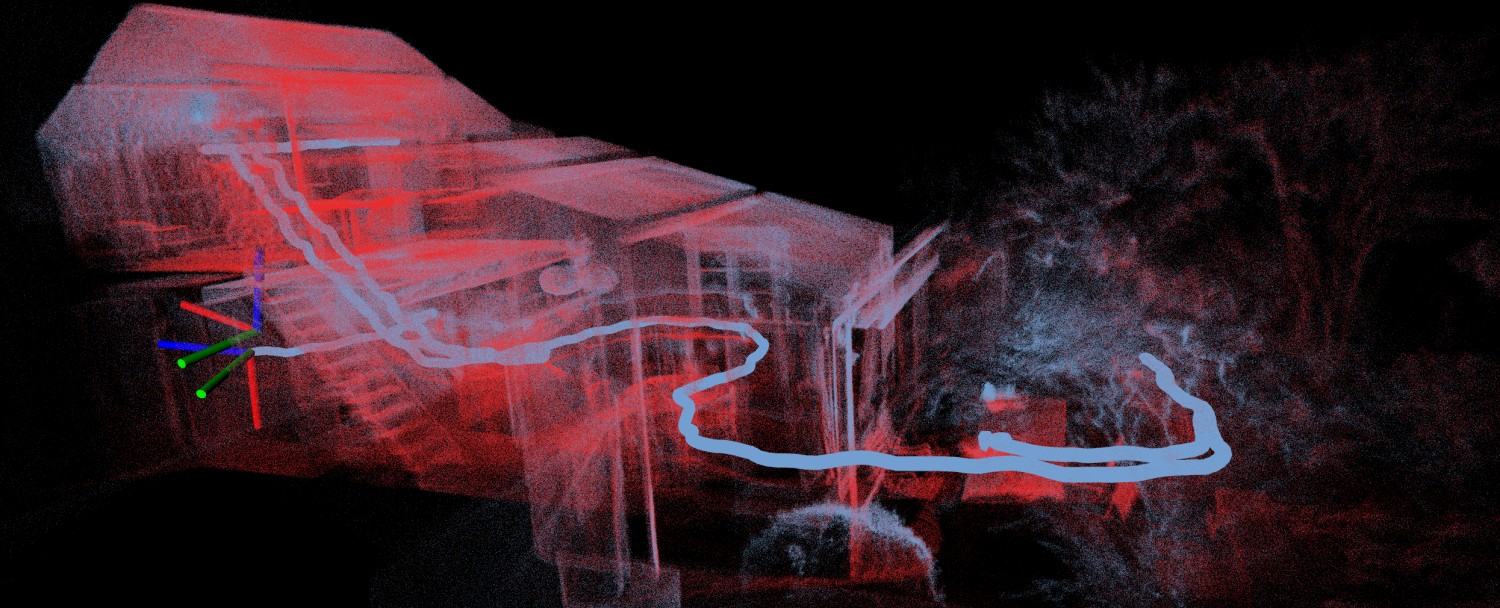} }}\\
    \subfloat[\centering Ongoing mapping experiment.]{{\includegraphics[width=.46\textwidth]{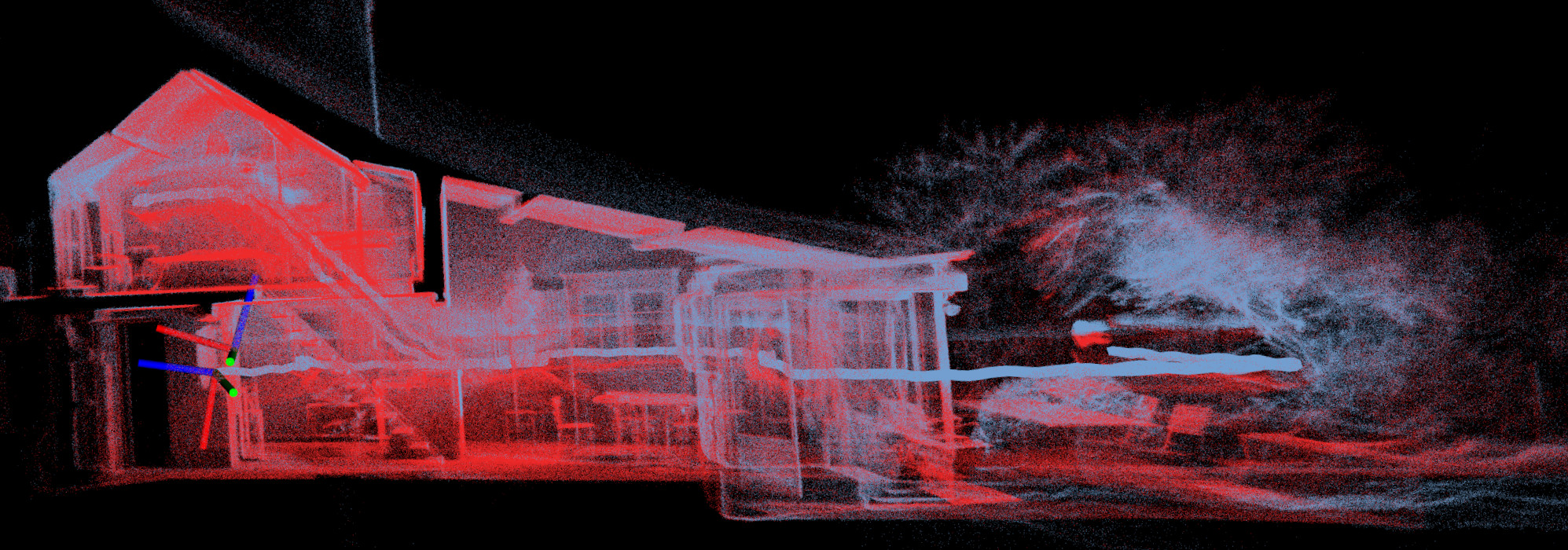} }}\\%
    \caption{Experiment D: dual non-rigid LiDAR setup. LiDARs are denoted L1 and L2. The blue and red colors indicate which LiDAR has observed this point.}%
    \label{fig:exp:dual-mid360}%
\end{figure}
The sensor setup and experimental results are shown in Figure \ref{fig:exp:mid360oak-color}. First, we notice the global accuracy of the map (shown only qualitatively in these figures), mainly thanks to the overall improved quality of the sensor setup. In fact, we now use more powerful and precise LiDAR data, together with wide angle global shutter cameras, in a dual forward-backward stereo manner. Here, the point density is sufficient, but we still notice that the LiDAR sensor placement might not be ideal. Despite our effort to tilt the LiDAR, we notice that we could potentially get better point clouds if the LiDAR was not mounted at such a height (on top of the operator's head).

\subsection{Dual non-rigid LIO with body-worn sensors}
We implement this system to try and improve the LIO sub-system of the previous experiment. The idea is to discard the helmet-mounted LiDAR and leverage i) the small weight of the Livox Mid-360 and ii) the abundance of velcro pads on jackets meant for security, tactical or emergency services. Here, we mount two LiDAR-inertial devices, simply fixed with velcro, on a military jacket taken from the Belgian Defence Clothing System\footnote{BDCS, \url{https://bdcs.be/}}. We choose to put the sensors on the chest, tilted 90$^\circ$ forward, and on the shoulder. Of course, in this configuration both sensors are not rigidly attached to each other, which differs from most multi-LiDAR solutions and adds complexity to the system. However, given our practical use cases, the increased flexibility of a non-rigid, dual-LiDAR setup presents a potential advantage that is worth the experiment. In terms of algorithm, our system uses a variant of Faster-LIO \cite{Bai2022FasterLIOLT}, in which we have implemented two odometry pipelines which share the same map representation. Although relatively simple, this system already shows promising results: we have a very complete map and the odometry stability is improved compared to a single LiDAR system. The sensor setup and experiments are shown in Figure \ref{fig:exp:dual-mid360}. However, we have had failure cases (complete divergence) where one of the LiDARs could not gather sufficient geometrical features to properly estimate its state. This happened mainly with the chest-mounted LiDAR, as its position is less suitable than the shoulder-mounted LiDAR for capturing good, long-lasting geometrical features. In the future, we will look into the implementation of additional constraints between both LiDARs so that the odometry can be improved.

\section{CONCLUSION}

In this paper, we explored the development of wearable 3D mapping systems for first responders. We evaluated different sensor configurations mounted on helmets or vests. Our experiments included a helmet-mounted Intel Realsense L515 ToF camera (limited by range and field of view), a Microsoft Azure Kinect (demonstrating improved performance but with limitations in range and use in outdoor scenarios), and a helmet-mounted LiDAR system with a camera (showing promise, but with potential limitations due to sensor placement). Finally, we investigated a non-rigid body-worn dual LiDAR setup with initial positive results. Our findings suggest that wearable 3D mapping systems have the potential to revolutionize emergency response capabilities. Future work will focus on refining sensor placement and exploring additional data fusion techniques to create even more robust and informative 3D maps for real-world scenarios.

\bibliographystyle{IEEEtran}
% \bibliography{refs}

\end{document}